\documentclass[letterpaper]{article} 
\usepackage{aaai2026}  
\usepackage{times}  
\usepackage{helvet}  
\usepackage{courier}  
\usepackage[hyphens]{url}  
\usepackage{graphicx} 
\usepackage{natbib}  
\usepackage{caption} 
\usepackage{algorithm}
\usepackage{algorithmic}

\usepackage{subfigure}
\usepackage{booktabs}
\usepackage{makecell}
\usepackage{arydshln}
\usepackage{multirow}
\usepackage{amssymb}
\usepackage{pifont}
\usepackage{diagbox}

\usepackage{newfloat}
\usepackage{listings}
\DeclareCaptionStyle{ruled}{labelfont=normalfont,labelsep=colon,strut=off} 
\floatstyle{ruled}
\newfloat{listing}{tb}{lst}{}
\floatname{listing}{Listing}
\title{RPTS: Tree-Structured Reasoning Process Scoring for Faithful Multimodal Evaluation}
\author{
    Haofeng Wang\textsuperscript{\rm 1},
    Yu Zhang\textsuperscript{\rm 1}\thanks{Corresponding author}
}
\affiliations{
    \textsuperscript{\rm 1}Harbin Institute of Technology, Harbin, China\\
    hfwang@ir.hit.edu.cn, zhangyu@ir.hit.edu.cn
}

\usepackage{bibentry}

\begin{document}

\maketitle

\begin{abstract}
Large Vision-Language Models (LVLMs) excel in multimodal reasoning and have shown impressive performance on various multimodal benchmarks. However, most of these benchmarks evaluate models primarily through multiple-choice or short-answer formats, which do not take the reasoning process into account. Although some benchmarks assess the reasoning process, their methods are often overly simplistic and only examine reasoning when answers are incorrect. This approach overlooks scenarios where flawed reasoning leads to correct answers. In addition, these benchmarks do not consider the impact of intermodal relationships on reasoning. To address this issue, we propose the Reasoning Process Tree Score (RPTS), a tree structure-based metric to assess reasoning processes. Specifically, we organize the reasoning steps into a reasoning tree and leverage its hierarchical information to assign weighted faithfulness scores to each reasoning step. By dynamically adjusting these weights, RPTS not only evaluates the overall correctness of the reasoning, but also pinpoints where the model fails in the reasoning. To validate RPTS in real-world multimodal scenarios, we construct a new benchmark, RPTS-Eval, comprising 374 images and 390 reasoning instances. Each instance includes reliable visual-textual clues that serve as leaf nodes of the reasoning tree. Furthermore, we define three types of intermodal relationships to investigate how intermodal interactions influence the reasoning process. We evaluated representative LVLMs (e.g., GPT4o, Llava-Next), uncovering their limitations in multimodal reasoning and highlighting the differences between open-source and closed-source commercial LVLMs. We believe that this benchmark will contribute to the advancement of research in the field of multimodal reasoning.
\end{abstract}

\begin{links}
    \link{Code \& Datasets}{https://github.com/wang-hao-feng/RPTS}
    \link{Extended version}{https://arxiv.org/abs/2511.06899}
\end{links}

\section{Introduction}
Recent advances in multimodal foundation models have demonstrated increasingly sophisticated capabilities in the combination of visual and textual information ~\cite{gpt4}. However, as these models begin to assist in evidentiary reasoning tasks such as criminal case analysis - where establishing reliable connections between surveillance footage (visual modality), forensic reports (textual modality), and other evidence is crucial, and where conclusions must follow rigorous, verifiable reasoning chains - two critical questions emerge: 1. Can current evaluations distinguish between logically valid reasoning and coincidentally correct conclusions? 2. Do existing frameworks capture the non-linear, cross-modal reasoning required to resolve conflicting evidence?

\begin{figure*}[htb]
    \centering
    \includegraphics[width=\linewidth]{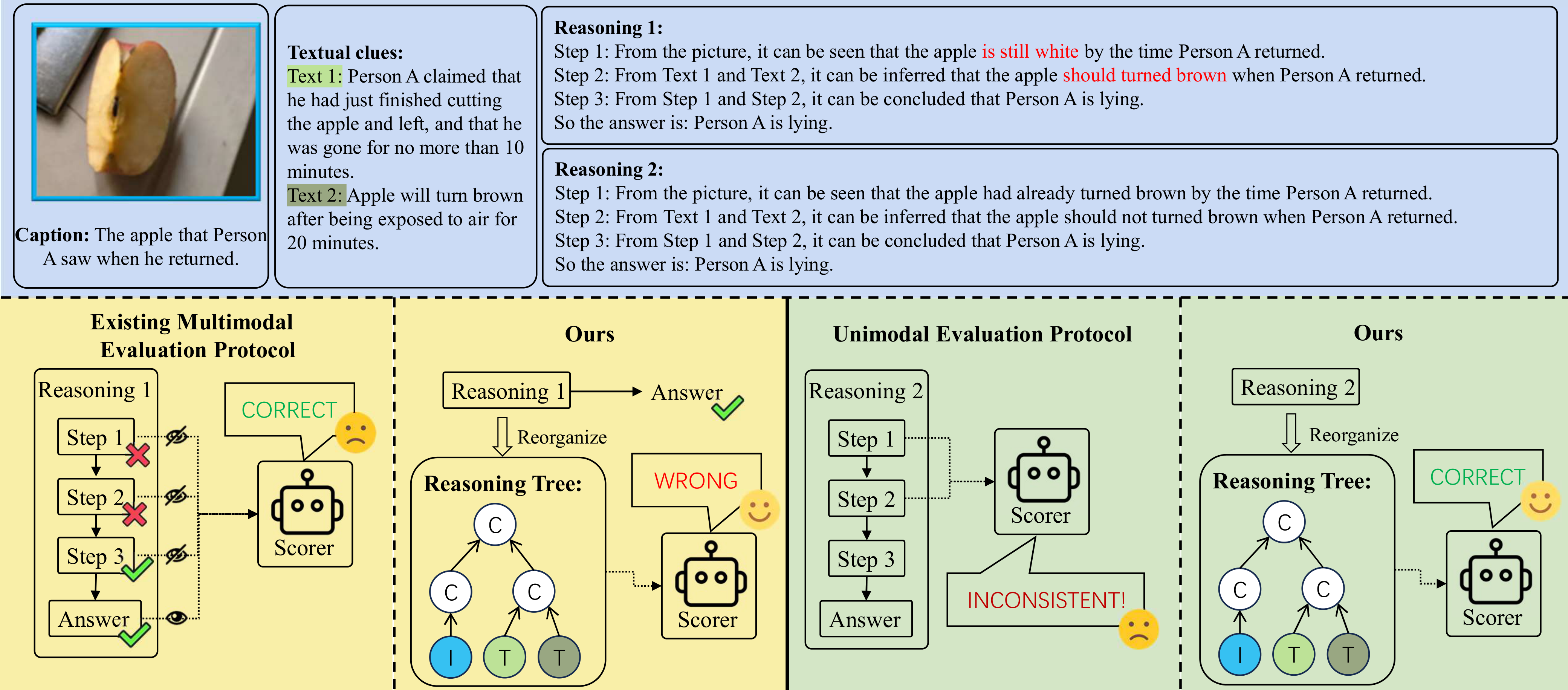}
    \caption{Comparison between Unimodal, existing Multimodal benchmarks and our RPTS. \textbf{Left}: Current multimodal benchmarks fail to detect instances where reasoning errors are present, yet the answer remains correct. \textbf{Right}: The unimodal approach is unable to handle reasoning involving conflicting information across different modalities.}
    \label{compare}
\end{figure*}

Most existing benchmarks focus solely on task accuracy through multiple choice or short answer formats ~\cite{mmvet, mmmu}, completely ignoring the reasoning process. This approach fails to detect when models arrive at correct conclusions through flawed reasoning, a phenomenon we call "right answers for wrong reasons" , as illustrated on the left side of Figure \ref{compare}. The few works that examine reasoning processes ~\cite{roscoe, receval} typically adopt an oversimplified linear evaluation framework. These approaches are fundamentally mismatched to the complex, non-linear nature of real-world reasoning, where multimodal evidence may appear conflicting yet collectively support valid conclusions.
To address these limitations, we introduce a novel evaluation metric: Reasoning Process Tree Score (RPTS), designed to assess multimodal reasoning processes.  The core innovation of RPTS lies in its tree-structured representation of reasoning, where leaf nodes correspond to atomic evidence units (visual or textual) and non-leaf nodes capture the hierarchical derivation of intermediate conclusions. This structure inherently accommodates the non-linear interactions characteristic of multimodal reasoning. Furthermore, RPTS incorporates two key hyperparameters whose adjustable values enable precise quantification of both global and local logical consistency, thereby facilitating accurate error localization within the reasoning chain.

\begin{figure}[htb]
    \centering
    \includegraphics[width=\linewidth]{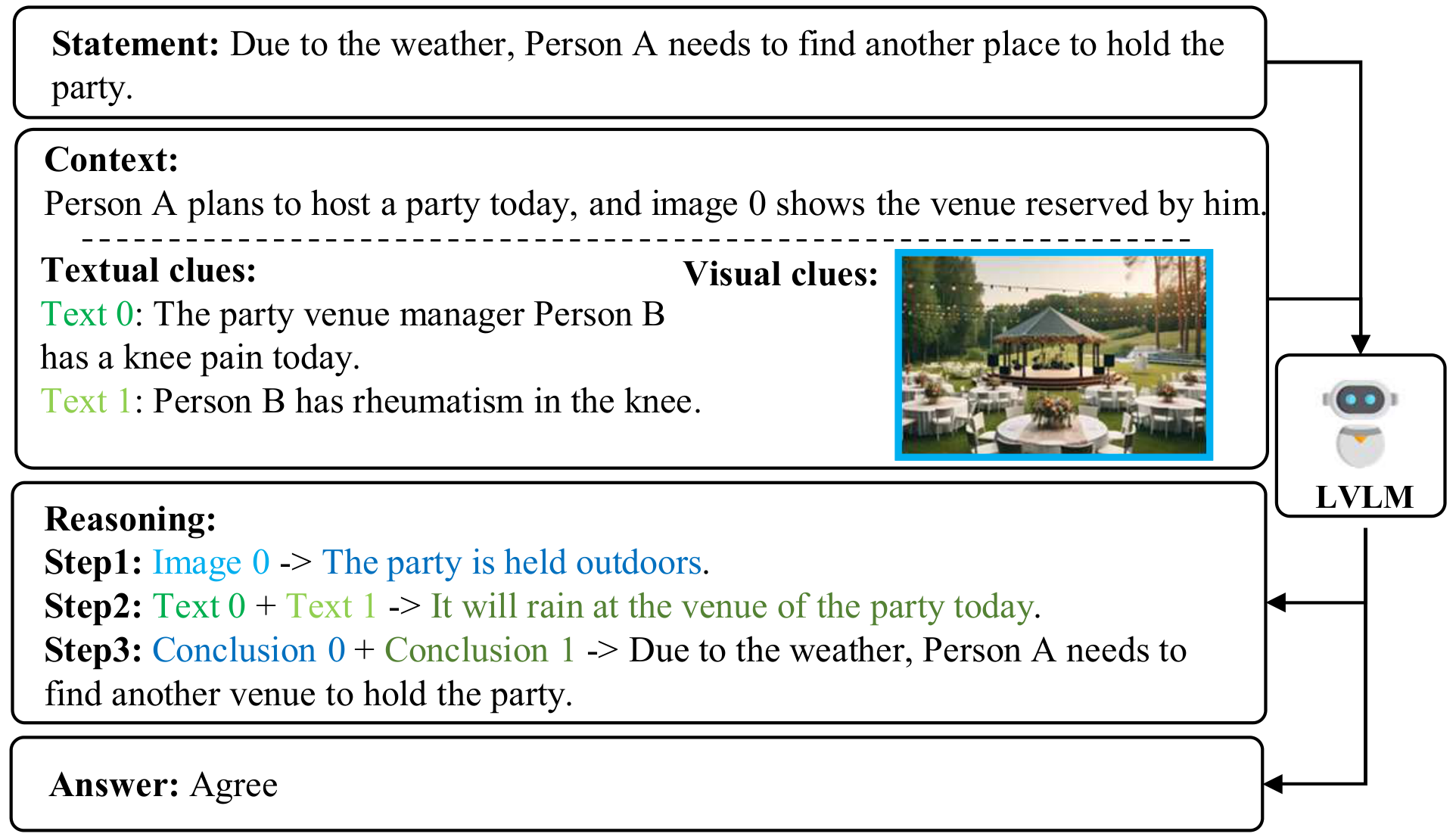}
    \caption{An example of RPTS-Eval.}
    \label{example}
\end{figure}

To support comprehensive evaluation using RPTS, we developed the RPTS-Eval benchmark comprising 390 carefully constructed reasoning instances. Figure \ref{example} shows examples of RPTS-Eval. Each instance contains complete and reliable multimodal atomic evidence for building reasoning trees. To systematically investigate how inter-modal relationships affect reasoning, we defined three distinct modality interaction types: guided (related without interference), adversarial (related with interference), and independent (unrelated), with each instance manually annotated accordingly. Experimental results on RPTS-Eval demonstrate RPTS's capability to identify flawed reasoning and precisely localize errors, while revealing significant limitations in current LVLMs' reasoning abilities. The primary contributions of our work can be summarized as follows:
\begin{itemize}
    \item We introduce a new metric, RPTS, for detecting correct conclusions based on faulty reasoning and genuinely logical reasoning processes, reflecting both overall and local logic of reasoning, achieving error localization.
    \item We constructed RPTS-Eval, a novel benchmark for multimodal reasoning evaluation. Compared to existing datasets, RPTS-Eval provides reliable multimodal annotations of atomic evidence that facilitate a rigorous assessment of reasoning processes.
    \item We define three types of relationships between modalities in reasoning, which clarify the classification of multimodal reasoning.
    \item We conducted extensive experiments with our RPTS-Eval. The results reveal that current open-source LVLMs have difficulty drawing conclusions from images for further inference and show varying performance across different languages.
\end{itemize}

\section{Related Work}
\subsubsection{MLLM Evaluation Benchmarks}
Classic multimodal benchmarks typically assess the specific reasoning abilities of the models. For example, OK-VQA\cite{okvqa} evaluates a model's capacity to leverage external knowledge for reasoning, while VCR\cite{vcr} focuses on human-related common sense reasoning. To evaluate the comprehensive capabilities of a model, researchers have proposed various benchmarks, such as MMBench\cite{mmbench}, SEED-Bench\cite{seedbench}, MM-Vet\cite{mmvet}, and MMMU\cite{mmmu}. These benchmarks scrutinize the reasoning abilities of models from diverse perspectives, often employing multiple choice or simplified formats to facilitate the evaluation process. InfiMM-Eval\cite{infimm-eval} incorporates the reasoning process into the evaluation, scoring the entire reasoning process. However, it cannot perform a more detailed analysis of reasoning and its evaluation method cannot exclude cases where incorrect reasoning leads to a correct answer.
\subsubsection{Verify Reasoning Process}
Recent studies have introduced various techniques for evaluating reasoning processes. ROSCOE\cite{roscoe} proposes a set of quality metrics to assess reasoning from four perspectives: semantic alignment, semantic similarity, logical correctness, and semantic coherence. ReCEval \cite{receval} evaluates reasoning based on two criteria: whether the reasoning steps are correct and whether new information is derived from the reasoning. REVEAL provides a dataset to validate whether a model can be used to verify the reasoning process. 

\begin{figure}[htb]
    \centering
    \includegraphics[width=\linewidth]{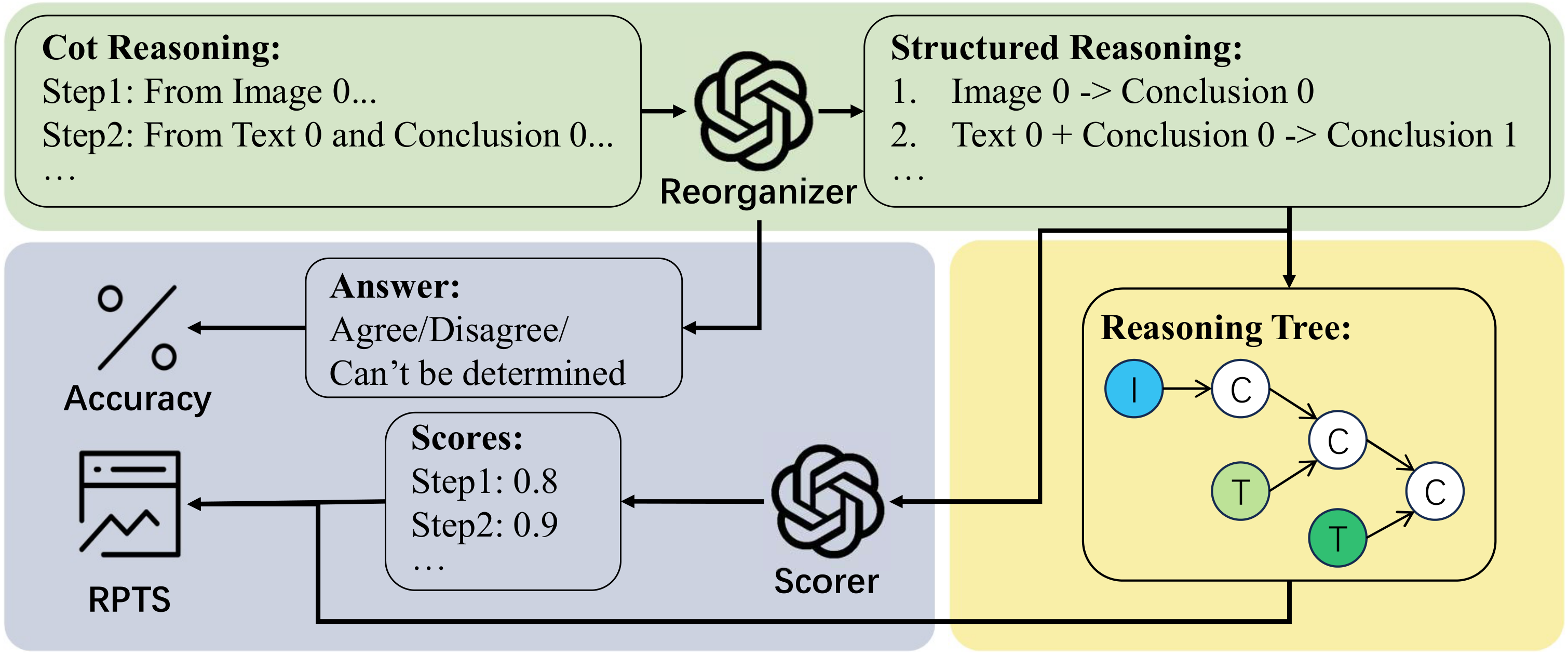}
    \caption{The calculation process of RPTS and accuracy.}
    \label{evaluation process}
\end{figure}

\section{RPTS}
The computation of RPTS consists of two stages: Reasoning Parsing and Metric Calculation. Figure \ref{evaluation process} illustrates the workflow of RPTS alongside the accuracy computation.

\subsection{Reasoning Parsing}
To construct the reasoning tree, we first parse the model's reasoning into a structured format: "[PREMISE] + [PREMISE] + ... $\rightarrow$ [CONCLUSION]", where '[PREMISE]' can be derived from visual clues, textual clues, or intermediate conclusions from prior steps. However, existing open-source MLLMs cannot strictly adhere to this output format. To address this, we first employ chain-of-thought (CoT) prompting to guide the model in generating step-by-step reasoning with explicit premises. Subsequently, we use GPT-4 to reformat the reasoning into a structured, easily parsable representation. The parsed reasoning results can also be utilized for accuracy computation.

\subsection{LLM-Based Scorer}
Now, each reasoning step in our approach strictly adheres to the "[PREMISE] + [PREMISE] + ... $\rightarrow$ [CONCLUSION]" format. Prior studies \cite{llm-evaluator, G-Eval, gptscore, touchstone, visit-bench, mmvet, infimm-eval} have demonstrated LLMs' effectiveness in assessing model reasoning. Therefore, we utilize a LLM to score reasoning, but with a unique twist: we only evaluate individual reasoning steps, not the entire process. This method allows for more precise evaluations by preventing the influence of other reasoning elements on the scores. Before we input the reasoning into scorer, we first preprocess the model's reasoning by eliminating redundant text clues, merging conclusions from images, substituting unnumbered texts and conclusions with all relevant clues and conclusions, and removing reasoning without '[PERMISE]'.
For scoring reasoning according with image, we calculate the semantic similarity of conclusions directly derived from images against the ground truth. For other reasoning, we input the premises and conclusion into LLM to assess their logical coherence. The score given by scorer ranges from 0 to 1, with higher scores indicating stronger logical reasoning. However, as illustrated in Figure \ref{compare}, there are instances where the model’s selected premises may not directly support the given conclusion, though they may be justified within the broader reasoning context. To address this, if the initial score is below 0.5, we re-evaluate using all text clues and previously derived conclusions as new premises, and then applying a 0.8 penalty for incorrect premises. We select the higher of the two scores as the final assessment.

\begin{figure}[htb]
    \centering
    \includegraphics[width=\linewidth]{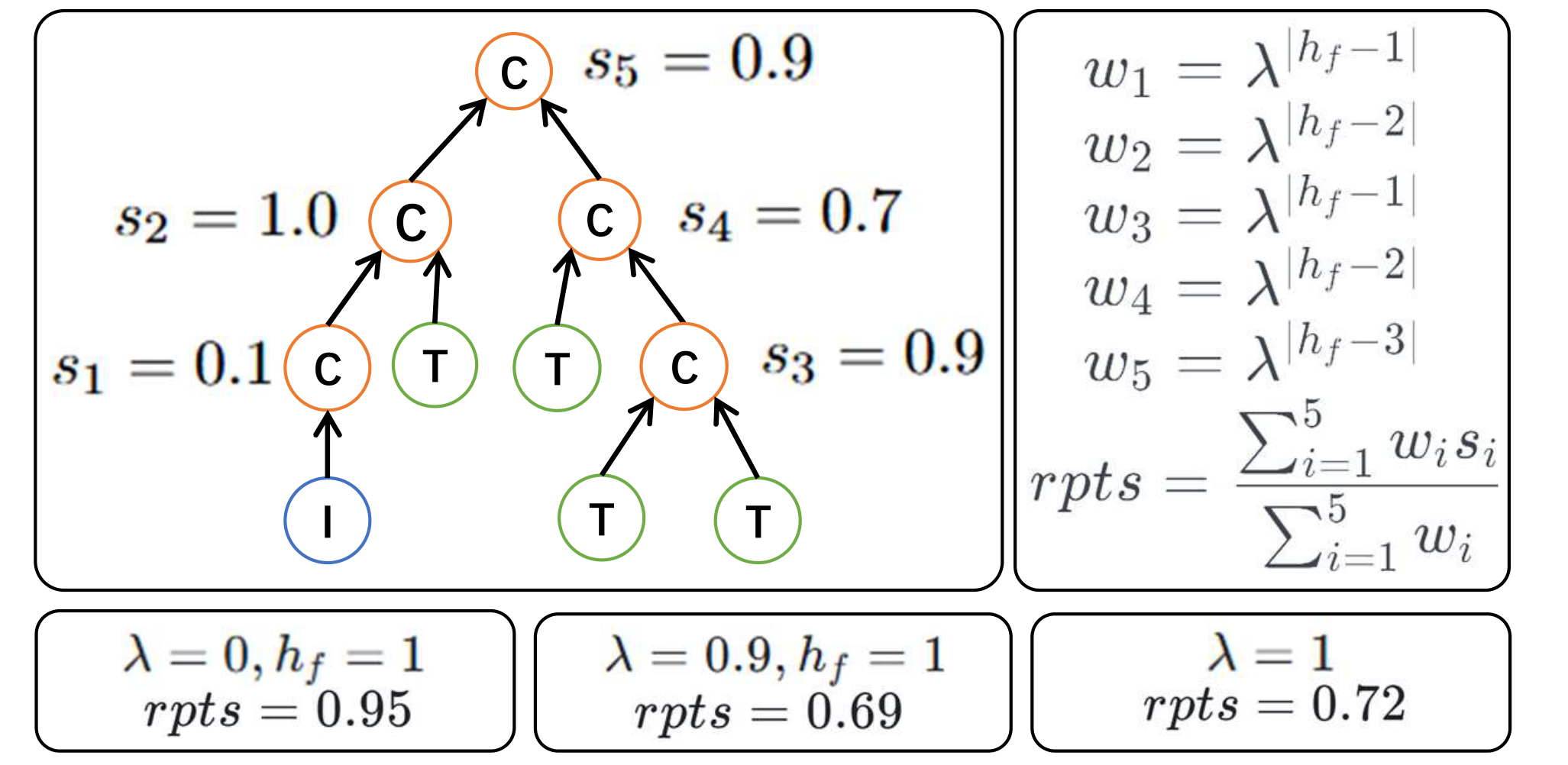}
    \caption{Examples of different hyperparameter settings for RPTS. C, I and T respectively represent conclusion, visual clue and textual clue.}
    \label{rpts example}
\end{figure}

\begin{figure*}[htb]
    \centering
    \includegraphics[width=\linewidth]{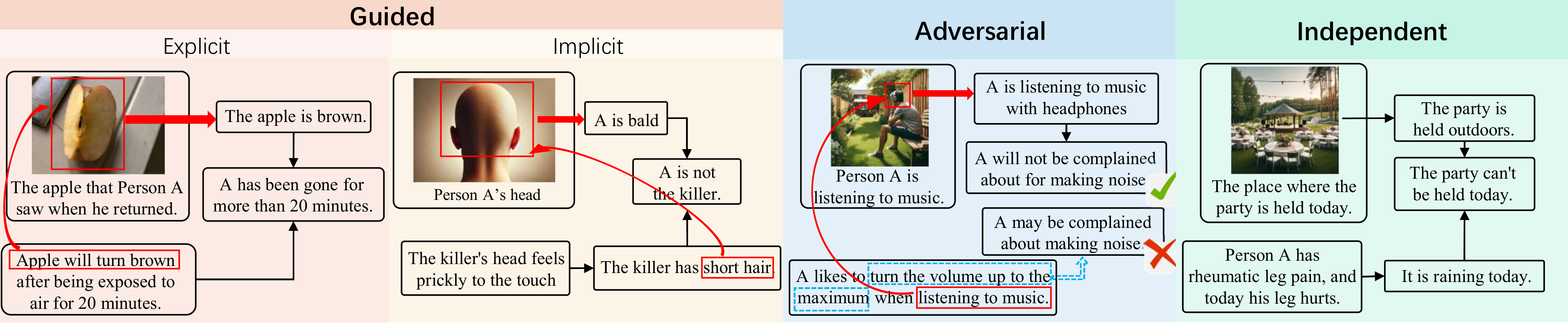}
    \caption{Three types of relationships between modalities. The filled arrow represents the relation between image and text, and the outlined arrow represents the interference between them.}
    \label{classification example}
\end{figure*}

\subsection{Reasoning Process Tree Score}
Considering the unique structure of reasoning, we can model the process as an reasoning tree, as depicted in Figure \ref{rpts example}. In this tree, the leaf nodes represent context, visual clues and textual clues, while the non-leaf nodes correspond to individual steps of inference. This tree, alongside parameters $\lambda$ and $h_f$, is used to weight each inferential step. The weight assigned to $n_i$ is defined as 
\begin{equation}
    \label{eqn-1}
    w_i = \lambda^{|h_f - h|}
\end{equation} where $n_i$ is the node corresponding to the $i^{th}$ step of inference, $h$ denotes the height of $n_i$, defined as the number of edges on the longest path from $n_i$ to any leaf node. $h_f$ is an integer that indicates the step most focused on by the RPTS, meaning that the weight is maximized at step $h_f$ , with the weights decaying along the reasoning tree centered around this step. $\lambda$ is the decay factor, which controls the speed at which the reasoning weight decays. The overall score of the reasoning tree, RPTS, is calculated as 
\begin{equation}
    \label{eqn-2}
    RPTS = \frac{\sum^N_{i=1}w_is_i}{\sum^N_{i=1}w_i}
\end{equation} where $N$ is the number of steps in the inference process, and $s_i$ is the score of the $i^{th}$ inferential step. By adjusting $\lambda$ and $h_f$, we can finely tune the emphasis on global versus local aspects of the inference process.

Figure \ref{rpts example} illustrates the scoring outcomes under three different settings of $\lambda$ and $h_f$. If we aim to assign equal importance to each reasoning step, we can set $\lambda$ to 1, as shown in the top-right corner of Figure \ref{rpts example}. In this case, RPTS represents the average score of all reasoning steps. Conversely, if we only wish to focus on the first step, i.e., reasoning at height 1, as depicted in the bottom-left corner, we set $\lambda$ to 0 and $h_f$ to 1. Under this configuration, only the score of the first step contributes to the RPTS computation. In the main experiment presented in Table \ref{main results}, the hyperparameters we configured are consistent with those shown in the bottom-right corner of Figure \ref{rpts example}. We focused more on the first step in reasoning, so we set $h_f = 1$. As shown in Table \ref{tab:benchmark statistic}, most of the reasoning trees in RPTS-Eval have a height of no more than 7, and we want the weight of each reasoning step to be no less than 0.5 when calculating the RPTS.  Based on Formula \ref{eqn-1}, we can calculate $\lambda > \sqrt[|h_f-7|]{0.5} \approx 0.891$. Therefore, we selected $\lambda = 0.9$.

\section{RPTS-Eval}
To enable a fine-grained analysis of models' reasoning capabilities, we propose RPTS-Eval, a novel multimodal reasoning evaluation benchmark. Each reasoning instance in RPTS-Eval contains reliable visual-textual clues that serve as leaf nodes of the reasoning tree, facilitating structured reasoning assessment.
\subsection{Data Collection}
We aim to developing a high-quality multimodal reasoning evaluation benchmark, using a meticulously designed methodology to assess model reasoning performance. Each sample in RPTS-Eval can be viewed as a multimodal reasoning story. Constructing such stories automatically poses significant challenges, even GPT-4 struggles to generate reasoning stories with sufficiently coherent logic. In addition, it is difficult to find suitable stories from online sources, and the time investment required for manually designing stories is substantial. To address these issues, the process of constructing data can be broadly divided into the following steps:
\subsubsection{Collating Inspiration.}
To reduce the difficulty of manually designing stories, we use GPT-4 to assist annotators. First, we ask an annotator to design a few reasoning stories and input them into GPT-4 as examples. Following the approach of MM-Vet\cite{mmvet}, we then require GPT-4 to generate reasoning stories encompassing six types of capabilities, based on the given examples. However, we define two distinct capabilities that differ from MM-Vet: Image Comparison (IC) and Spatial Awareness (SA). The remaining four capabilities—Recognition (Rec), OCR, Commonsense Reasoning (Com), and Math—are identical to those defined in MM-Vet. The specific definitions of IC and SA are as follows:
\begin{itemize}
    \item \textbf{Image Comparison(IC):} The model compares two images to spot similarities or differences. This is a basic human skill, as we learn a lot about the world by observing and comparing things.
    \item \textbf{Spatial Awareness(SA):} This covers spatial skills, like recognizing fixed positions or understanding how objects relate to each other from different viewpoints.
\end{itemize}
As mentioned above, the stories generated by GPT-4 lack logical consistency. Therefore, annotators only draw inspiration from these stories rather than using them directly, thereby reducing the difficulty of story design. For example, GPT-4 generates a story set on a rainy day, and our annotators draw inspiration from this, such as the idea of 'rain,' and design reasoning tasks based on that inspiration. For instance, they might infer whether the box is open or closed based on whether there’s water inside, or they might reason about whether outdoor activities can continue based on certain pre-rain features.

\subsubsection{Constructing Data}
This phase involves two annotators, each assigned to different reasoning stories. First, the annotators need to design two reasoning paths based on the stories. These two reasoning paths should use similar clues to arrive at opposite conclusions. Then, the annotators should design statements, contexts, visual and textual clues, reasoning steps, and required abilities for the data based on the reasoning paths. Finally, the annotators need to find suitable images according to their design. The images for RPTS-Eval are sourced from the internet and text-to-image modals.

\begin{table*}[htb]
    \centering

    \begin{tabular}{ccccccc}
    \toprule
       Benchmark & Size & Images & Answer Format & Metric & Evaluate Reasoning \\
    \midrule
        MMMU\cite{mmmu}  & 11.5K & 12.5K & Option/Open Answer & Accuracy & \ding{55} \\
       MM-Vet\cite{mmvet}  & 218 & 200 & Shot answer & GPT4-score & \ding{55} \\
       InifMM-Eval\cite{infimm-eval}  & 279 & 342 & Reasoning & GPT4-score & \ding{51} \\
    \midrule
       RPTS-Eval(Ours)  & 390 & 374 & Reasoning & Accuracy+RPTS & \ding{51} \\
    \bottomrule
    \end{tabular}

    \caption{The comparison between RPTS-Eval and other existing benchmarks.}
    \label{tab:compair}
\end{table*}

\subsection{Quality Control}
To ensure data quality, each piece of data is validated by two validators. We reference InifMM-Eval\cite{infimm-eval} and conduct a comprehensive evaluation of the data based on the following criteria:
\begin{itemize}
    \item \textbf{Logical Scoring:} Check how statements, context, visuals, and reasoning connect, and score them to ensure strong logic.
    \item \textbf{Multimodality:} Remove samples that don't require both visual and text clues for reasoning (single-modality solvable).
    \item \textbf{Subjectivity and Discrepancy Check:} Discard or edit overly subjective data or cases where validator reasoning clashes with ground truth.
    \item \textbf{Missing or Redundant abilities:} Validators flag missing or unnecessary annotated reasoning abilities.
\end{itemize}
We excluded data where there was disagreement between the two validators as well as those with low logical scores.

\subsubsection{Multimodal Reasoning Classification.}
To better investigate the reasoning capabilities of multimodal models, we categorize the constructed data into three types based on the relationships between modalities during reasoning. Examples of these three reasoning types are illustrated in Figure \ref{classification example}.
\begin{itemize}
    \item \textbf{Guided}: By utilizing information from one modality, it becomes possible to determine which information should be retrieved from another modality to complete the reasoning process. The relationships between modalities are categorized into two types: explicit and implicit. Explicit relationships are defined as cases where one modality directly indicates the information that needs to be obtained from another modality. In contrast, implicit relationships involve cues from one modality that require reasoning to infer which information should be retrieved from the other modality.
    \item \textbf{Adversarial}: In some cases, one modality can negatively influence information extraction from another, either by leading to irrelevant/incorrect data or by preventing useful information from being retrieved at all.
    \item \textbf{Independent}: Modalities don’t influence each other, information must be gathered separately from each for reasoning.
\end{itemize}


\begin{table}[]
    \centering
    \begin{tabular}{cccc}
    \toprule
       \textbf{Statistics} & \textbf{Percentage} & \textbf{Statictic} & \textbf{Percentage}\\
    \midrule
        \multicolumn{4}{c}{Capabilities}\\
        Rec & 83.08\% & Math & 24.87\% \\
        Com & 40.00\% & OCR & 18.46\% \\
        SA & 28.97\% & IC & 5.13\% \\
        \midrule
        \multicolumn{4}{c}{Answer}\\
        agree & 50.00\%  & disagree & 50.00\% \\
        \midrule
        \multicolumn{4}{c}{Relationship} \\
        Guided & 84.62\% & Adversarial & 6.92\% \\
        Independent & 8.46\% \\
        \midrule
        \multicolumn{2}{c|}{Reasoning steps} & \multicolumn{2}{c}{Reasoning tree height}\\
        $\leq 2$ & \multicolumn{1}{c|}{3.85\%} & $\leq 2$ & 0.51\% \\ 
        $3$ & \multicolumn{1}{c|}{42.82\%} & $3$ & 11.03\% \\
        $4$ & \multicolumn{1}{c|}{32.56\%} & $4$ & 52.56\% \\
        $5$ & \multicolumn{1}{c|}{13.08\%} & $5$ & 26.92\% \\
        $\geq 6$ & \multicolumn{1}{c|}{7.69\%} & $\geq 6$ & 8.67\% \\        
    \bottomrule
    \end{tabular}
    \caption{Key statistics of the RPTS-Eval benchmark. As each reasoning instance need one or more capabilities, the sum of percentage is larger than 100\%.}
    \label{tab:benchmark statistic}
\end{table}

\subsection{Dataset Statistics}
In summary, our RPTS-Eval benchmark comprises 390 inferences linked to a total of 374 images. Table \ref{tab:benchmark statistic} depicts the distribution across multiple dimensions of RPTS-Eval. Since most tasks require the recognition of objects in images, object recognition capability plays a dominant role. Given that the data is constructed with paired answers, the two types of answers in RPTS-Eval are evenly distributed, which helps mitigate the effects of model bias. The relationships between modalities are primarily based on guided, as the reasoning for the last two types are more challenging to construct. The majority of inferences can be made within 5 steps, and when the inference is represented as a tree, the tree height is typically below 6. For a comparison with other benchmarks, please refer to Table \ref{tab:compair}.

\begin{table*}[htb]
    \centering
    \begin{tabular}{c|ccc|ccc}
    \toprule
          \multirow{2}{*}{\textbf{Models}} & \multicolumn{3}{c|}{\textbf{English}} & \multicolumn{3}{c|}{\textbf{Chinese}}\\ \cline{2-4} \cline{5-7}
           & \textbf{Acc} & \textbf{RPTS}$\uparrow$ & \textbf{Acc$_{filtered}$} & \textbf{Acc} & \textbf{RPTS}$\uparrow$ & \textbf{Acc$_{filtered}$}\\
    \midrule
        Llava-v1.5-7B & 0.64 & 0.63 & 0.48(-0.16) & 0.35 & 0.57 & 0.24(-0.12) \\
        Llava-Next-7B & 0.62 & 0.47 & 0.32({-0.29}) & 0.13 & 0.41 & 0.06(-0.07) \\
        Qwen-VL-Chat & 0.57 & 0.61 & 0.41(-0.16) & 0.39 & 0.61 & 0.25(-0.14) \\
        ShareGPT4V-7B & 0.58 & 0.56 & 0.38(-0.20) & 0.34 & 0.50 & 0.19(-0.15) \\
        InternVL2-8B & 0.63 & 0.67 & 0.53(-0.10) & 0.46 & 0.66 & 0.37(-0.08) \\
        \hdashline
        Llama-3.2-11B & 0.68 & 0.68 & 0.56(-0.12) & 0.41 & 0.63 & 0.29(-0.12) \\
        InstructBLIP & 0.56 & 0.59 & 0.41(-0.16) & - & - & - \\
        Llava-v1.5-13B & 0.56 & 0.59 & 0.41(-0.15) & 0.41 & 0.58 & 0.28(-0.13) \\
        Llava-Next-13B & 0.62 & 0.51 & 0.34(-0.27) & 0.23 & 0.46 & 0.11(-0.12) \\
        ShareGPT4V-13B & 0.59 & 0.50 & 0.32(-0.27) & 0.35 & 0.58 & 0.26(-0.09) \\
        \hdashline
        InternVL2-26B & 0.65 & 0.70 & 0.55(-0.10) & 0.54 & 0.74 & 0.45(-0.08) \\
        \hdashline
        Llava-Next-34B & 0.68 & 0.71 & 0.60(-0.08) & 0.46 & 0.68 & 0.37(-0.09) \\
        InternVL2-40B & {0.74}* & {0.76}* & {0.67}*({-0.06})* & {0.57}* & 0.75 & {0.52}(-0.05)* \\
        \hdashline
        InternVL2-76B & 0.73 & \underline{0.79} & \underline{0.70}(\underline{-0.04}) & \underline{0.60} & \underline{0.77} & \underline{0.57}(\underline{-0.03}) \\
        Llama-3.2-90B & \underline{0.79} & 0.67 & 0.66(-0.12) & 0.56 & {0.77}* & {0.52}*({-0.04})* \\
        \hdashline
        GPT-4o & \textbf{0.86} & \textbf{0.84} & \textbf{0.84}(\textbf{-0.02}) & \textbf{0.72} & \textbf{0.86} & \textbf{0.70}(\textbf{-0.02}) \\
        
    \bottomrule
    \end{tabular}
    \caption{Results of different models on RPTS-Eval with cot prompt. We set $\lambda = 0.9$, $h_f = 1$ when calculate RPTS. For each column, the highest, the second, and the third highest figures are highlighted by \textbf{bold}, \underline{underline} and {star}*. \textbf{Acc}: Accuracy.}
    \label{main results}
\end{table*}

\section{Experiments}
\subsection{Models and Evaluation Metrics}
To validate the challenging nature of RPTS-Eval and the capability of the RPTS evaluation metric analysis model, we conducted experiments in both Chinese and English across various models. The open-source models tested include InstructBLIP\cite{instructblip}, InternVL2\cite{intervl2}, ShareGPT4V\cite{sharegpt4v}, Llava-v1.5\cite{llava1.5}, Llava-Next\cite{llavanext} and Qwen-VL-Chat\cite{qwenvl}, detailed in Appendix A; the sole close-source model examined is GPT-4o. We evaluate the reasoning ability of the model by combining accuracy and RPTS, and analyze the problems of the model.

\subsection{Scorer Selection}
To select an appropriate scoring model, we randomly sampled 200 reasoning instances and manually scored them. Concurrently, we selected five distinct models of varying types and sizes as potential scorers, evaluating their performance against human-assigned scores. Table \ref{tab:scorers} presents the \textbf{M}ean \textbf{A}bsolute \textbf{E}rror (MSE) between the scores generated by these models and those assigned by humans. Based on the minimal discrepancy observed, we opted for GPT-4 as our designated scoring model.

\begin{table}[]
    \centering
    \begin{tabular}{cc}
    \toprule
       Model & $\overline{\Delta}$ \\
    \midrule
        Qwen2-7B & 0.216 \\
        Llama-3-8B & 0.231 \\
        \hdashline
        Qwen2-72B & 0.143 \\
        Llama-3-70B & 0.152 \\
        \hdashline
        GPT-4 & \textbf{0.095} \\
    \bottomrule
    \end{tabular}
    \caption{Mean absolute error ($\bar{\Delta}$) between different LLM scores and human scores}
    \label{tab:scorers}
\end{table}

\subsection{Experiment Settings}
Our experiment involves both Chinese and English languages and performs chain-of-thought(COT)\cite{cot} reasoning on the RPTS-Eval benchmark. All tests were performed in a zero-shot setting using a greedy decoding strategy to assess the models' inferential abilities. To optimize the COT reasoning outcomes, we designed five Chinese prompts and seven English prompts, selecting the most effective one from each language for our experiments. All tests were carried out on an NVIDIA A100 GPU. When reasoning, we set the temperature of each model to 0 and use greedy decoding.

\begin{figure*}[htbp]
    \centering
    \includegraphics[width=0.7\linewidth]{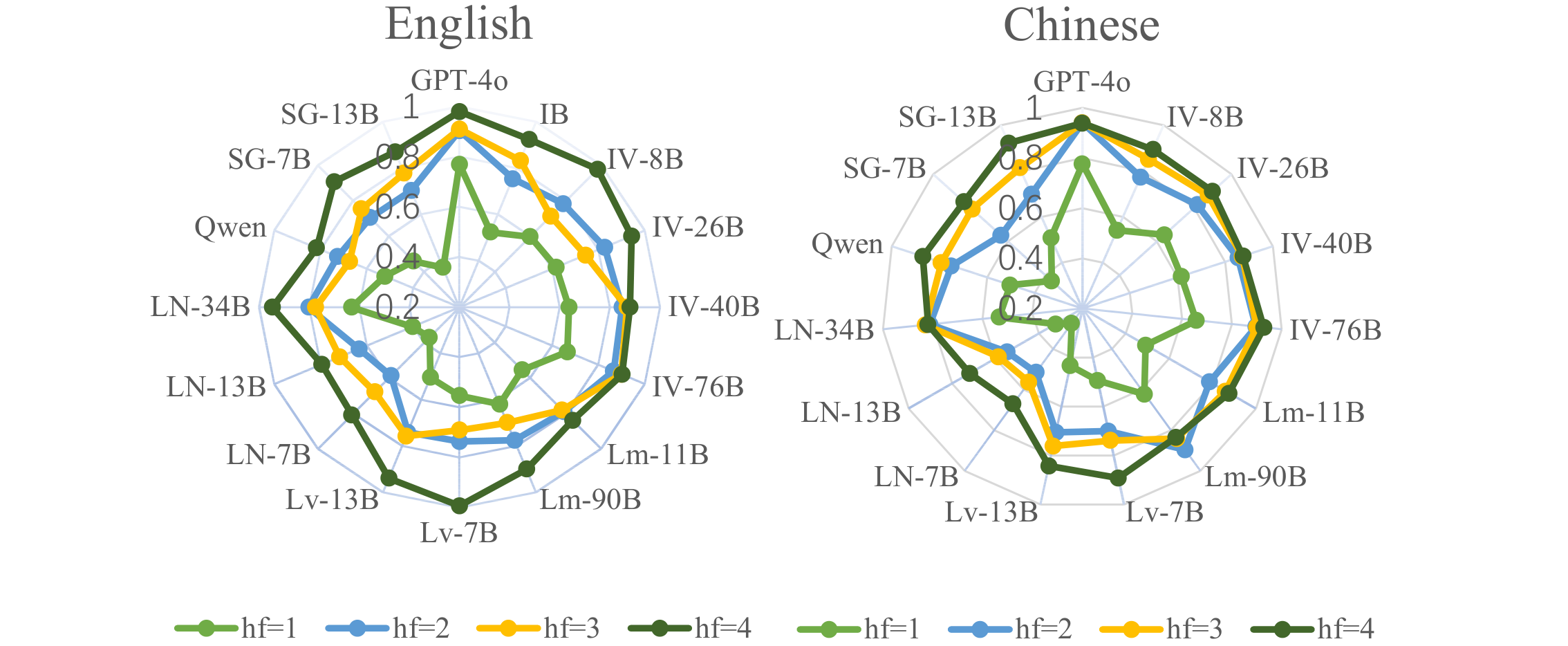}
    \caption{RPTS scores for $h_f \in \{1, 2, 3, 4\}$ and $\lambda = 0$. \textbf{IB}: InstructBLIP; \textbf{IV}: InternVL2; \textbf{Lv}: Llava-v1.5; \textbf{LN}: Llava-Next; \textbf{Qwen}: Qwen-VL-Chat; \textbf{SG}: ShareGPT4V;}
    \label{beta}
\end{figure*}

\subsection{Results and Analysis}
Table \ref{main results} presents model performance on RPTS-Eval. Beyond inference accuracy and mean RPTS scores, we applied an RPTS-based filter (score < 0.5) to exclude cases where correct conclusions arose from flawed reasoning. Results show all models experienced accuracy declines, with GPT-4o least affected—consistent with its stronger logical capacity. In the results of GPT-4, the lower RPTS scores are associated with erroneous reasoning and the model's failure to capture certain infomation. Conversely, the open-source models demonstrated a lack of logical robustness in their reasoning processes, leading to more pronounced decreases due to often generating irrelevant or illogical outputs. Despite these models' lower accuracy, their RPTS scores were not significantly impacted. We hypothesize that this is due to two primary reasons: 1. Disconnection between the inference outcomes and the intended targets. While the models initially could reason based on the specified targets, they gradually lost focus on the targets as the number of reasoning steps increased, resulting in conclusions that diverged from the intended data targets. 2. Recurrent generation of identical sentences. Across various sizes, the open-source models consistently produced repetitive reasoning that, while logically sound, failed to reach the desired conclusions. These factors led to reduced accuracy but did not substantially affect the logical integrity of the inferences, as reflected in the relatively high RPTS scores. In addition, Appendix B shows the performance of the model on six capabilities.

\begin{table}[!htb]
    \centering
    \begin{tabular}{c|cc|cc}
    \toprule
        \multirow{2}{*}{\textbf{Models}} & \multicolumn{2}{c|}{\textbf{English}} & \multicolumn{2}{c}{\textbf{Chinese}} \\ \cline{2-3} \cline{4-5}
         & \textbf{V} & \textbf{T} & \textbf{V} & \textbf{T} \\
    \midrule
        InternVL2-8B & 0.50 & 0.76 & 0.54 & 0.94 \\
        Llava-1.5-7B & 0.42 & 0.78 & 0.40 & 0.67 \\
        Llava-Next-7B & 0.36 & 0.52 & 0.22 & 0.58 \\
        ShareGPT4V-7B & 0.35 & 0.62 & 0.30 & 0.70 \\
        \hdashline
        Llama-3.2-11B & 0.52 & 0.83 & 0.4 & 1.0 \\
        InstructBLIP & 0.40 & 0.69 & - & - \\
        Llava-1.5-13B & 0.41 & 0.66 & 0.37 & 0.78 \\
        Llava-Next-13B & 0.36 & 0.57 & 0.29 & 0.73 \\
        Qwen-VL-Chat & 0.45 & 0.74 & 0.45 & 0.66 \\
        ShareGPT4V-13B & 0.18 & 0.57 & 0.42 & 0.75 \\
        \hdashline
        InternVL2-26B & 0.53 & 0.80 & 0.57 & 0.84 \\
        \hdashline
        Llava-Next-34B & 0.54 & 0.80 & 0.52 & 0.72 \\
        InternVL2-40B & 0.61 & 0.90 & 0.60 & 0.88 \\
        \hdashline
        InternVL2-76B & 0.60 & 0.92 & 0.60 & 0.87 \\
        Llama-3.2-90B & 0.58 & 0.79 & 0.6 & 0.75 \\
        \hdashline
        GPT-4o & 0.72 & 0.88 & 0.75 & 0.96 \\
    \bottomrule
    \end{tabular}
    \caption{RPTS score for drawing conclusions from visual clues(\textbf{V}) or textual clues(\textbf{T}).}
    \label{first step score}
\end{table}

\begin{table}[htb]
    \centering
    \begin{tabular}{c|cccccccccc}
    \toprule
       \diagbox[]{$h_f$}{$\lambda$} & 0.2 & 0.4 & 0.6 & 0.8 & 1.0 \\
    \midrule
       1  & 0.647 & 0.671 & 0.690 & 0.703 & 0.713 \\
       2  & 0.768 & 0.743 & 0.728 & 0.719 & 0.713 \\
       3  & 0.733 & 0.733 & 0.727 & 0.720 & 0.713 \\
       4  & 0.733 & 0.734 & 0.729 & 0.721 & 0.713 \\
    \bottomrule
    \end{tabular}
    \begin{tabular}{c|cccccccccc}
    \toprule
       \diagbox[]{$h_f$}{$\lambda$} & 0.2 & 0.4 & 0.6& 0.8 & 1.0 \\
    \midrule
       1  & 18.21 & 16.15 & 14.36& 10.51 & 10.26 \\
       2  & 8.97 & 9.74 & 9.23 & 8.97 & 10.26 \\
       3  & 10.00 & 9.74 & 10.51 & 9.49 & 10.26 \\
       4  & 10.26 & 9.49 & 9.49 & 9.23 & 10.26 \\
    \bottomrule
    \end{tabular}
    \caption{Sensitivity analysis of RPTS: values (top) and percentage of low-score correct answers (bottom) under various $\lambda$ and $h_f$.}
    \label{tab:sensitivity analysis}
\end{table}

\subsubsection{Step Analysis.} To further identify the causes of errors in our model, we initiated an analysis from the perspective of inference steps. We set $\lambda = 0$ and varied $h_f$ at values of 1, 2, 3, and 4 to compute the average RPTS score. Figure \ref{beta} displays the relationship between RPTS scores and $h_f$ across two languages. As evident from the Figure \ref{beta}, with the exception of GPT-4o, RPTS scores at $h_f = 1$ are unsatisfactory across all models. This indicates that the models encounter issues at the initial inference step, where conclusions are drawn directly from the visual and textual clues, leading to subsequent errors in reasoning. To further explore the specific causes, we calculated the average RPTS scores derived separately from visual and textual clues. The results, as shown in Table \ref{first step score}, reveal that open-source models still lack sufficient capabilities in image processing. They fail to derive necessary information from images for subsequent reasoning tasks based on specific inferential questions.

\subsubsection{Sensitivity Analysis}
To further investigate the impact of different $\lambda$ and $h_f$ values on RPTS and the correctness of reasoning, we conducted a sensitivity analysis using the reasoning results from InternVL-26B. Table \ref{tab:sensitivity analysis} presents the RPTS values and the proportion of filtered reasoning paths for various $\lambda$ and $h_f$ settings. From the table, it can be observed that small variations in $\lambda$ do not significantly alter the RPTS values or the proportion of filtered reasoning paths. However, changes in $h_f$ lead to notable differences, particularly when $\lambda$ is small. This aligns with our design intention: a smaller $\lambda$ reduces the influence of non-$h_f$ steps, thereby making RPTS more closely reflect the score of the $h_f$ step.

\section{Conclusion}
In this paper, we introduce RPTS-Eval, a benchmark specifically designed to meticulously examine the reasoning processes of models. We also define three types of relationships between modalities in multimodal reasoning. Furthermore, we propose a new metric, RPTS, aimed at addressing issues where incorrect reasoning still results in correct outcomes, thereby facilitating a detailed analysis of model reasoning. Our results indicate that current open-source Large Visual Language Models struggle to derive necessary conclusions from images for subsequent reasoning. We also observed a significant disparity in the capabilities of models between Chinese and English contexts, suggesting that existing training methodologies fall short in transferring multimodal abilities from English to other languages.

\section{Acknowledgments}
We sincerely thank the anonymous reviewers for their insightful comments and suggestions. This work was supported by the National Natural Science Foundation of China
(No. 62476066).

\bibliography{aaai2026}

\end{document}